\documentclass{article}
\usepackage{graphicx} 

\usepackage{amsmath, amssymb}
\usepackage{mathtools}
\mathtoolsset{showonlyrefs}

\usepackage{caption}
\usepackage{subcaption}
\usepackage{url}
\usepackage{pgfplots}
\pgfplotsset{compat=1.18}

\usepackage[style=authoryear, natbib=true, sorting=none]{biblatex}
\usepackage{algorithm}
\usepackage{algpseudocode}
\algtext*{EndFor}

\usepackage[hidelinks]{hyperref}
\usepackage{stmaryrd}

\usepackage{booktabs}

\newcommand{\EE}{\boldsymbol{E}}
\newcommand{\ZZ}{\boldsymbol{Z}}
\newcommand{\XX}{\boldsymbol{X}}
\newcommand{\YY}{\boldsymbol{Y}}

\title{Betting on Moments: Legendre Jumper Martingales for Online Exchangeability Testing}
\author{
        Johan Hallberg Szabadv\'ary\\
        \textit{Department of Mathematics, Stockholm University}\\
        \texttt{johan.hallberg.szabadvary@math.su.se}
        }
\date{May 2026}

\usepackage{amsthm}
\newtheorem{remark}{Remark}

\begin{document}

\maketitle

\begin{abstract}
    A fundamental assumption in statistics and machine learning is that ``the future looks like the past,'' formalized as exchangeability: the joint data distribution is order-invariant. In practice, this assumption is often violated due to distribution shifts over time. Early detection of exchangeability violations is crucial to prevent performance degradation and enable timely interventions like model retraining. Conformal test martingales offer a flexible, distribution-free framework for sequential exchangeability testing with guaranteed false-alarm rate control by betting against the uniformity of conformal p-values. While alternatives such as plug-in martingales and mixture-based strategies exist, computationally efficient baselines like the Simple Jumper are limited to detecting mean location shifts. We propose a family of conformal test martingales based on shifted Legendre polynomials that extend the Simple Jumper to higher-order moments. The Simple Legendre Jumper replaces linear betting functions with polynomials of arbitrary degree, enabling rapid detection of variance, skewness, and other higher-order deviations. The Product Legendre Jumper combines multiple polynomial degrees into a single betting function but suffers from exponential state-space growth, termed the jumping tax. To resolve this, we introduce the Variational Legendre Jumper, which employs a mean-field approximation to reduce complexity to constant time per step with minimal power loss, providing an expressive, scalable framework for real-time distribution shift monitoring.
\end{abstract}

\textbf{Keywords:}
Conformal test martingales, Concept drift detection, Sequential exchangeability testing, Shifted Legendre polynomials, Variational mean-field approximation

\section{Introduction}
A standard assumption in statistics and machine learning (ML) is that the underlying data-generating distribution remains stable over time, that is, ``the future looks like the past.'' This can be formalised as the \emph{exchangeability assumption}, which posits that the joint distribution of the data sequence is invariant under permutations. Exchangeability is more general than the \emph{IID assumption}, in which data are generated independently from the same probability distribution, but the two are closely related and even equivalent under some conditions (see below). Violations of exchangeability---due to distribution shifts, concept drift, or other non-stationarities---can compromise the validity and predictive performance of ML models, which motivates the need for reliable online testing procedures that detect such violations promptly while controlling false-alarm rates.

Classical fixed-sample tests are ill-suited for this sequential setting because they either inflate type I error when applied repeatedly or require strong parametric assumptions. \emph{Conformal test martingales} (CTMs) \citep{alrw2} provide a flexible, distribution-free framework for sequentially testing exchangeability by monitoring p-values derived from conformal prediction methods. Under exchangeability, these p-values are independent and uniformly distributed on $[0,1]$. Intuitively, a CTM describes the capital process of a player who gambles against the hypothesis that the p-values $p_1,p_2,\dots$ are uniformly and independently distributed (which they are under exchangeability) such that the betting game is fair (ensured by the martingale property) and that the gambler never risks bankruptcy (see Section \ref{sec:preliminaries} for more details). If the p-values are independent and uniformly distributed, the gambler is not likely to gain much capital, but if they deviate from uniformity or exhibit dependency, she could become very rich. This intuitive picture allows us to informally describe a simple online test of exchangeability: let the player gamble against the p-values. If she accumulates sufficiently large capital, which is unlikely under the exchangeability hypothesis, we discard the exchangeability hypothesis. The rationale behind this test rests on \emph{Cournot's principle}, often considered to be the only bridge between probability theory and its application \citep{cournot1843exposition}, although rarely referenced by name. The principle states that if the probability of an event is very small, one can be practically certain that it will not occur. In our case, the probability that the gambler accumulates a large capital is unlikely if the exchangeability hypothesis is true; therefore, we can be practically certain that it will not happen under exchangeability. Because CTMs are martingales under the exchangeability assumption, the false alarm rate is controlled by Ville's inequality \citep{ville1939etude}, which quantifies ``large'' and ``unlikely'' (see Section \ref{sec:preliminaries}).

The connection between exchangeability and IID is established by de Finetti's representation theorem \citep{de1937prevision}: any exchangeable probability measure on $\mathbb{R}^{\infty}$ is a mixture of product measures. \citet[Section 7]{hewitt1955symmetric} extended this result to replace $\mathbb{R}$ with any Borel space and pointed out that, to the best of their knowledge, every measurable space known to have importance in applied science is Borel. Therefore, in almost every practically conceivable situation, testing the IID assumption is equivalent to testing the exchangeability assumption over an infinite horizon. In a finite horizon, when we have $N$ observations, the equivalence breaks down, as randomly permuting the dataset ensures exchangeability but not the IID assumption \citep{vovk2021retrain}. Nevertheless, a CTM can still be used to detect violations of the IID assumption in the finite horizon: a dataset that violates exchangeability must also violate the IID assumption.

Among CTMs, the Simple Jumper martingale \citep{alrw2} stands out for its computational efficiency and robust performance. By employing a Markov transition mechanism over a discrete grid of linear betting functions, it adapts to shifting environments by betting on location shifts—changes in the mean of the p-values. However, many practical distributional shifts affect higher-order moments such as variance, skewness, or kurtosis, which the Simple Jumper may fail to detect.

To overcome this limitation, we generalise the Simple Jumper by employing shifted Legendre polynomials, an orthogonal basis on the unit interval known from classical smooth goodness-of-fit testing \citep{neyman1937smooth}, as betting functions. This leads to the \emph{Simple Legendre Jumper}, which replaces the linear betting function with polynomials of arbitrary degree to target deviations in specific moments beyond the mean.
Building on this, we introduce the \emph{Product Legendre Jumper}, which combines multiple polynomial degrees multiplicatively to simultaneously capture complex, multimodal, and asymmetric deviations from uniformity. While this enhances detection power, it introduces a computational bottleneck due to the exponential growth of the state space, a phenomenon we term the ``jumping tax.''
To address this, we develop the \emph{Variational Legendre Jumper}, which factorises joint adaptation into independent marginal components via a mean-field approximation inspired by variational inference. This reduces the computational complexity from exponential to linear in the number of polynomial degrees while retaining statistical power.
Finally, recognising that the optimal adaptation rate depends on unknown shift timescales, we propose the \emph{Composite Legendre Jumper}---a generalisation of the Composite Jumper \citep{alrw2}---which mixes over multiple jumping rates to provide robustness and a guaranteed wealth floor under exchangeability.
This suite of Legendre-based martingales offers a flexible, assumption-lean framework for reliable online exchangeability testing, with practical applicability in machine learning and related fields that require sequential decision-making under distributional shifts.

The remainder of this paper is organised as follows. Section~\ref{sec:preliminaries} reviews the Simple Jumper martingale and its connection to the first-degree shifted Legendre polynomial. Section~\ref{sec:legendre-polynomials} introduces Legendre polynomials and their shifted variants, including the historical context via Neyman’s smooth test. Section~\ref{sec:slj} defines the Simple Legendre Jumper for an arbitrary polynomial degree. Section~\ref{sec:plj} develops the Product Legendre Jumper and introduces the Variational Legendre Jumper as a scalable alternative. Section~\ref{sec:composite} defines the Composite Legendre Jumper, which mixes over several jumping rates to provide robustness to rate misspecification. Section~\ref{sec:state-space} establishes the validity of the standard grid for all degrees. Section~\ref{sec:empirical} presents the empirical results of a real-world classification task, and Section~\ref{sec:conclusion} offers concluding remarks. An open-source implementation of the Legendre Jumper martingales introduced in this paper, including the Variational and Composite architectures, is publicly available in the Python package \texttt{online-cp}\footnote{Source code and documentation are available on GitHub at \url{https://github.com/egonmedhatten/online-cp} and via PyPI at \url{https://pypi.org/project/online-cp}.} (starting from version 0.3.0) to facilitate reproducible research and practical applications.

\section{Preliminaries}\label{sec:preliminaries}
Let $\XX$ and $\YY$ be two measurable spaces, called the \emph{object space} and the \emph{label space}, respectively. Their Cartesian product $\ZZ:=\XX\times\YY$ is called the \emph{example space}. Each example $z = (x,y)\in\ZZ$ consists of an object $x$ and its associated label $y$. Our aim is to test whether a data sequence $z_1,z_2,\dots$ satisfies the exchangeability assumption. If we find enough evidence to reject the exchangeability hypothesis, we want to stop collecting data and raise a warning as quickly as possible.
A CTM is determined by a \emph{betting martingale} and a \emph{nonconformity measure}. To describe a CTM, we must first understand its parts. Let us begin with nonconformity measures.

A \emph{nonconformity measure} is a measurable function $A : \ZZ^{(*)}\times\ZZ\to\overline{\mathbb{R}}$, where $\ZZ^{(*)}$ is the set of all bags (or multisets) of elements of $\ZZ$. The purpose of a nonconformity measure is to score how ``strange'' or ``unusual'' an example $z\in\ZZ$ appears in comparison to the elements of a bag. It will always be somewhat debatable whether a particular function $A$ is suitable to determine nonconformity. Indeed, even a constant function satisfies the definition of a nonconformity measure, but will not be very useful to numerically score ``strangeness''. Nevertheless, any nonconformity measure, together with some auxiliary randomness determines a \emph{conformal transducer} $f:(\XX\times[0,1]\times\YY)^*\to[0,1]$. The output of a conformal transducer is a \emph{p-value}, defined as
\begin{equation}\label{eq:p-value}
	\begin{aligned}
		p_n := f(&x_1,\tau_1,y_1, \dots,x_n,\tau_n,y_n)                                                                          \\
		                         & = \tfrac{|\{i=1,\dots,n:\alpha_i > \alpha_n\}| + \tau_n|\{i=1,\dots,n:\alpha_i=\alpha_n\}|}{n},
	\end{aligned}
\end{equation}
where $\alpha_i:=A(\lbag z_1,\dots,z_n\rbag,z_i)$ and the numbers $\tau_1,\tau_2,\dots$ are generated independently (of everything) from the uniform distribution on $[0,1]$. The fundamental theorem of conformal prediction \citep[Theorem 11.1]{alrw2} states that if $z_1,z_2,\dots$ are generated from an exchangeable probability distribution, the p-values $p_1,p_2,\dots$ are independent and distributed according to the uniform distribution on $[0,1]$.

The second component of a CTM is a \emph{betting martingale}. We describe how to construct them using \emph{betting functions} $f:[0,1]\to[0,\infty]$ (we allow them to take value $\infty$) satisfying $\int_0^1f(u)du = 1$. A \emph{betting martingale} is a measurable function $F:[0,1]^* \to [0,\infty]$ such that $F(\square) = 1$ (where $\square$ is the empty sequence) and for each sequence $(u_1,\dots,u_{n-1})\in[0,1]^{n-1}$, we have
\begin{equation}\label{eq:betting-martingale-prop}
	\int_0^1 F(u_1,\dots,u_{n-1},u)du = F(u_1,\dots,u_{n-1}).
\end{equation}
A convenient way to construct betting martingales is to choose, before each step $n$, a betting function $f_n$ that may depend on $u_1,\dots,u_{n-1}$ in a measurable manner, and construct
\begin{equation}\label{eq:betting-martingale}
	F(u_1,\dots,u_n) = \prod_{i=1}^nf_i(u_i).
\end{equation}
Clearly, the construction in \eqref{eq:betting-martingale} satisfies the condition \eqref{eq:betting-martingale-prop}, since,
\begin{equation}\label{eq:martingale-property}
    \begin{aligned}
	   \int_0^1& F(u_1,\dots,u_{n-1},u)du \\&
       = \prod_{i=1}^{n-1}f_i(u_i)\int_0^1f_n(u)du = F(u_1,\dots,u_{n-1}).
    \end{aligned}
\end{equation}
The process $(F_n = F(U_1,\dots,U_n): n \geq 0)$, is a martingale with respect to the natural filtration $\mathcal{F}_n = \sigma(U_1, \dots, U_n)$ generated by the sequence $(U_i: i \geq 1)$, provided that the $U_i$ are independent and uniformly distributed on $[0,1]$. It is clear that each $F_n$ is $\mathcal{F}_n$-measurable, since it depends only on $U_1,\dots,U_n$. Moreover, because each betting function $f_i$ is non-negative and integrates to one, $\mathbb{E}[|F_n|]\leq\infty$ for all $n$. Finally, \eqref{eq:martingale-property} can be equivalently expressed as
\begin{equation}
	\begin{aligned}
		\mathbb{E}[F_n \mid \mathcal{F}_{n-1}] & = \mathbb{E}\left[F_{n-1} \cdot f_n(U_n) \mid \mathcal{F}_{n-1}\right]             \\
		                                       & = F_{n-1} \cdot \mathbb{E}[f_n(U_n)] \\
                                               &= F_{n-1} \cdot \int_0^1 f_n(u) du = F_{n-1}.
	\end{aligned}
\end{equation}
Together, these properties define a martingale (see e.g. \citet{Williams_1991}). Because it starts from the value 1 and the betting functions are non-negative, the betting martingale itself is non-negative; it is therefore a \emph{test martingale}.

A \emph{conformal test martingale} is what is obtained when a betting martingale processes conformal p-values. Specifically,
\begin{equation}
	S_n = F(p_1,\dots,p_n),\quad n=0,1,\dots,
\end{equation}
where $p_1,p_2,\dots$ are computed by \eqref{eq:p-value} using some nonconformity measure $A$ is a CTM. We say that the CTM $S$ \emph{is determined by} $A$ and $F$. Conformal test martingales are automatically martingales, since
\begin{equation}
	\mathbb{E}[S_n\mid S_1,\dots,S_{n-1}] = S_{n-1}
\end{equation}
for all $n\geq1$, provided that the examples $z_1,z_2,\dots$ are generated from an exchangeable probability distribution on $\ZZ$.

This construction corresponds to a classical non-negative conformal martingale operating on the filtration generated by the conformal p-values. Recent advances have highlighted that such martingales are a special case of more general \emph{safe e-processes} that provide powerful and valid sequential tests by operating on reduced filtrations \citep{ramdas2022testingexchangeability}. Surprisingly, when considering the filtration generated by the full data sequence, any martingale adapted to this filtration under exchangeability is almost surely constant, offering no power for testing. This motivates the use of the reduced filtration generated by the conformal p-values, which is strictly coarser and allows construction of non-trivial martingales that can accumulate evidence against exchangeability

The \emph{Ville procedure} for testing exchangeability online consists in computing p-values using a conformal transducer determined by some nonconformity measure $A$ and run them through a CTM determined by $A$ and a betting martingale $F$. By Ville's inequality \citep{ville1939etude},
\begin{equation}
	\mathbf{P}\bigl(\exists\, n : S_n \geq C\bigr) \leq 1/C \quad \text{for all $C \geq 1$},
\end{equation}
where $\mathbf{P}$ denotes any exchangeable distribution of examples. Therefore, the event $\{S_n \geq 1/\alpha\}$ constitutes level-$\alpha$ evidence against exchangeability at any stopping time. Thus, if we raise an alarm when $S_n\geq 1/\alpha$, the false-alarm rate is guaranteed not to exceed $\alpha$. This property is what guarantees the validity of CTMs. If, for example, we raise an alarm---perhaps triggering retraining of the ML model---when $S_n$ exceeds 100, the probability of ever raising a false alarm is at most $1\%$.

This paper focuses exclusively on the construction and analysis of the betting martingale component of a CTM, taking the nonconformity measure $A$ as a given function beyond our consideration. The choice and design of the nonconformity measure $A$, which determines the conformal p-values, is an important complementary aspect addressed in other works, including \cite{hore2026conformalchangepointlocalization}, which introduced the conformal Neyman–Pearson lemma for optimal detection of distributional changes. Going forward, we will sometimes use CTM and betting martingale interchangeably, always with the understanding that the CTM is obtained by letting the betting martingale bet on conformal p-values.

The art of constructing powerful CTMs lies in choosing the betting functions $f_i$ adaptively: a well-chosen bet grows wealth rapidly when the exchangeability assumption is violated, whereas the integral constraint in \eqref{eq:betting-martingale-prop} ensures that no wealth is gained on average under exchangeability. The remainder of this section reviews the Simple Jumper martingale, which provides a concrete and computationally efficient mechanism for this adaptive choice.

The Simple Jumper (SJ) martingale \citep{alrw2} builds on the betting functions
\begin{equation}\label{eq:SJ-betting-function}
	f_{\tilde{\epsilon}}(p) = 1 + \tilde{\epsilon}(p-1/2),\quad p\in[0,1]
\end{equation}
where $\tilde{\epsilon}\in \EE := \{-1,-1/2,0,1/2,1\}$. Additionally, it depends on the \emph{jumping rate} $J\in(0,1]$. For any probability measure $\mu$ on $\EE^{\infty}$, the function
\begin{equation}\label{eq:SJ-betting-martingale}
	F(p_1,\dots,p_n) := \int\prod_{i=1}^nf_{\tilde{\epsilon}}(p_i)\mu(d(\tilde{\epsilon}_1,\tilde{\epsilon}_2, \dots))
\end{equation}
is a CTM. For the Simple Jumper, the measure $\mu$ is defined as the probability distribution of a Markov chain with state space $\EE$, defined by the initial state $\widetilde{\varepsilon}$ being chosen from the uniform distribution on $\EE$. The transition function maintains the same state with probability $1-J$, and chooses a new state from the uniform probability measure on $\EE$ with probability $J$. Note that the CTM \eqref{eq:SJ-betting-martingale} is a deterministic function despite the Markov chain being stochastic. The Simple Jumper martingale can be computed using Algorithm \ref{alg:SJ}.
\begin{algorithm}[ht]
	\caption{Simple Jumper betting martingale}
	\label{alg:SJ}
	\begin{algorithmic}
		\State \textbf{Parameters:} state space $\EE$ and jumping rate $J\in(0,1]$.
		\State \textbf{Input:} p-values $p_1,p_2,\dots$.
		\State \textbf{Output:} martingale $S_1,S_2,\dots$..
		\State $C_{\tilde{\epsilon}} \gets 1/|\EE|$ for all $\tilde{\epsilon}\in\EE$
		\State $C \gets 1$
		\For{$n = 1, 2, \dots$}
		\For{$\tilde{\epsilon} \in \EE$:
			$C_{\tilde{\epsilon}} \gets (1-J) C_{\tilde{\epsilon}} + (J/|\EE|)C$}
		\EndFor
		\For{$\tilde{\epsilon} \in \EE$:
		$C_{\tilde{\epsilon}} \gets C_{\tilde{\epsilon}}f_{\tilde{\epsilon}}(p_n)$}
		\EndFor
		\State $S_n \gets C := \sum_{\tilde{\epsilon}\in\EE}C_{\tilde{\epsilon}}$
		\EndFor
	\end{algorithmic}
\end{algorithm}

\begin{remark}
	In practical implementations of conformal test martingales, repeated multiplication of betting functions can lead to numerical underflow or overflow due to the accumulation of very small or very large values. To mitigate this, it is advisable to perform the martingale computations in log space. Specifically, rather than updating the martingale capital ($S_n = \prod_{i=1}^n f_i(p_i)$) directly, one accumulates the sum of logarithms: $\log S_n = \sum_{i=1}^n \log f_i(p_i)$, and exponentiates only when the martingale value is needed for reporting (threshold comparison can of course also be carried out in log scale). This approach preserves the exact martingale property while significantly improving numerical stability.
\end{remark}

Our first observation is that, setting $\epsilon = \tilde{\epsilon}/2$, the betting function \eqref{eq:SJ-betting-function} can be written
\begin{equation}\label{eq:legendre-betting-function-k1}
	f_{\epsilon}(p) = 1 + \epsilon(2p-1) = 1 + \epsilon\widetilde{P}_1(p),
\end{equation}
where $\widetilde{P}_1(p)$ is the shifted Legendre polynomial of degree 1. The Legendre polynomials, named after the French mathematician Adrien-Marie Legendre, are a system of complete and orthogonal polynomials with a wide range of applications \citep{szeg1939orthogonal}. In this note, we will illustrate that they can profitably be applied as betting functions in betting martingales in various ways.

\section{Legendre polynomials}\label{sec:legendre-polynomials}
The Legendre polynomials can be defined in many ways, but the most straightforward one is by direct construction as an orthogonal system with respect to the weight function $w(x) = 1$ over the interval $[-1,1]$. In other words, $P_n(x)$ is a polynomial of degree $n$ such that
\begin{equation}
	\int_{-1}^1P_m(x)P_n(x)dx = 0\quad\text{if $n\neq m$.}
\end{equation}
With the additional condition that $P_n(1)=1$, this uniquely determines every Legendre polynomial: $P_0(x) = 1$ is the only possibility for degree 0. $P_1$ must be orthogonal to $P_0$, so that $P_1(x) = x$, and $P_2$ is determined by requiring orthogonality to $P_0$ and $P_1$, and so on. It is interesting to note that several other constructions are possible, e.g. by a recursive relationship, differential equations, and so on. Particularly compact is the Rodrigues formula:
\begin{equation}
	P_n(x) = \frac{1}{2^n n!}\frac{d^n}{dx^n}(x^2 - 1)^n.
\end{equation}
Importantly, Legendre polynomials have definite parity. That is, they are even or odd according to their degree. Specifically,
\begin{equation}\label{eq:parity}
	P_n(-x) = (-1)^nP_n(x).
\end{equation}
Because our focus lies on using the Legendre polynomials as betting functions, we limit our consideration to the \emph{shifted Legendre polynomials}, defined as
\begin{equation}\label{eq:shifted-LP}
	\widetilde{P}_n(x) = P_n(2x-1).
\end{equation}
The ``shifting function'' $x\mapsto2x-1$ is an affine transformation that bijectively maps the unit interval $[0,1]$ to $[-1,1]$, which implies that the shifted Legendre polynomials are orthogonal on $[0,1]$. In fact, one can show that
\begin{equation}
	\int_0^1\widetilde{P}_m(x)\widetilde{P}_n(x)dx = \frac{1}{2n+1}\delta_{mn},
\end{equation}
where $\delta_{mn}$ is the Kronecker delta (it is equal to zero if $m\neq n$, and one if $m=n$).
The first few shifted Legendre polynomials are listed in Table \ref{tab:shifted-LP}.
\begin{table}[ht]
	\centering
	\begin{tabular}{c|c}
		$n$ & $\widetilde{P}_n(x)$              \\
		\hline
		0   & 1                                 \\
		1   & $2x-1$                            \\
		2   & $6x^2 - 6x + 1$                   \\
		3   & $20x^3 - 30x^2 + 12 x - 1$        \\
		4   & $70x^4 - 140x^3 + 90x^2 -20x + 1$
	\end{tabular}
	\caption{The first few shifted Legendre polynomials.}
	\label{tab:shifted-LP}
\end{table}

Importantly for our constructions below, the shifted Legendre polynomials of degree $k\geq1$ trivially satisfy
\begin{equation}\label{eq:integral-Legendre}
	\int_0^1\widetilde{P}_k(x)dx = 0
\end{equation}
because they are all orthogonal to $\widetilde{P}_0(x) = 1$ by construction.

The application of Legendre polynomials to detect deviations from uniformity has a rich history in classical statistics. This history originated with Karl Pearson, who pioneered the problem of goodness-of-fit testing and devised the classical $\chi^2$ test \citep{pearson1900chi2}. Neyman's smooth test builds directly on this foundation: his paper \citet{neyman1937smooth} is explicitly dedicated to Pearson's memory\footnote{Karl Pearson died on April 27 1936}, acknowledging him as ``the originator of the problem of a test for goodness of fit and was first to advance its solution.'' Neyman argued that general omnibus tests, such as Pearson's $\chi^2$ test, lack statistical power against specific smooth departures from the null distribution. To construct a more powerful targeted test, Neyman built on the theoretical framework he had recently developed with Egon Pearson (Karl Pearson's son) on unbiased critical regions \citep{neyman1936contributions}. He modelled alternative probability densities as continuous distributions, smoothly deformed by a system of orthonormal polynomials, specifically, the normalised shifted Legendre polynomials. By truncating this polynomial series at a given degree $k$, Neyman's smooth test explicitly targets shifts in the first $k$ moments, such as location, variance, skewness, and kurtosis \citep{rayner1989smooth}.

Our sequential betting framework can be viewed as a game-theoretic and conformal analogue to Neyman's classical batch test. Neyman's smooth test models the alternative density as an exponential family $f(p;\boldsymbol{\theta}) \propto \exp\bigl(\sum_k \theta_k \widetilde{P}_k(p)\bigr)$, using shifted Legendre polynomials as sufficient statistics within a likelihood ratio framework. Our construction uses shifted Legendre polynomials directly as betting functions via $1 + \epsilon\widetilde{P}_k(p)$, which can be viewed as a first-order approximation of the exponential tilt. The connection is thus motivational rather than a formal generalisation: while Neyman used Legendre polynomials to construct static test statistics for fixed data samples, we leverage them as dynamic betting functions to sequentially compound martingale wealth. This historical connection naturally motivates our generalisations of the Simple Jumper: by incorporating higher-degree Legendre polynomials into the betting function, the martingale can systematically bet on the exact same smooth, multi-moment deviations from uniformity that Neyman originally identified.

\section{Simple Legendre Jumper}\label{sec:slj}
The first, and simplest generalisation of the Simple Jumper which, as we have seen, uses the betting function \eqref{eq:legendre-betting-function-k1}, is to use an arbitrary shifted Legendre polynomial in place of $\widetilde{P}_1$. Thus, the \emph{Simple Legendre Jumper} (SLJ) uses the betting function
\begin{equation}\label{eq:SimleLegendreJumper-bf}
	f_{\epsilon}^{(k)}(p) = 1 + \epsilon\widetilde{P}_k(p),\quad p\in[0,1],
\end{equation}
where $\epsilon\in\EE:=\{-1/2,-1/4,0,1/4,1/2\}$, and the same jumping rate parameter. Note that the Simple Legendre Jumper of order $k=1$ recovers the standard Simple Jumper martingale. The choice of $\EE$ is admittedly somewhat arbitrary, as remarked in \citet{alrw2}, but the dependence on $\EE$ is not heavy. We have simply transformed the standard grid in \citet{alrw2} to the scale of the shifted Legendre polynomials.

\begin{figure}[ht]
	\centering
	\begin{tikzpicture}[
			declare function={
					P1(\p) = 2*\p - 1;
					P2(\p) = 6*\p^2 - 6*\p + 1;
					P3(\p) = 20*\p^3 - 30*\p^2 + 12*\p - 1;
					P4(\p) = 70*\p^4 - 140*\p^3 + 90*\p^2 - 20*\p + 1;
					f(\p,\eps) = 1 + \eps * P1(\p);
				}
		]
		\begin{axis}[
				width=10cm,
				height=8cm,
				xmin=0, xmax=1,
				ymin=0, ymax=2.5,
				xlabel={p-value},
				ylabel={$f_\epsilon^{(1)}$},
				legend pos=north east,
				legend style={draw=black!30},
				legend cell align={left},
				domain=0:1,
				samples=100,
				thick 
			]

			\addplot [black, loosely dotted] {f(x, -1/2)};
			\addlegendentry{$\epsilon = -1/2$}

			\addplot [black!80, densely dotted] {f(x, -1/4)};
			\addlegendentry{$\epsilon = -1/4$}

			\addplot [black!60, densely dashed] {f(x, 0)};
			\addlegendentry{$\epsilon = 0$}

			\addplot [black!50, dashdotted] {f(x, 1/4)};
			\addlegendentry{$\epsilon = 1/4$}

			\addplot [black!40, dashed] {f(x, 1/2)};
			\addlegendentry{$\epsilon = 1/2$}

		\end{axis}
	\end{tikzpicture}
	\caption{Betting functions of the Simple Legendre Jumper for $k=1$.}
	\label{fig:betting-functions-k1}
\end{figure}
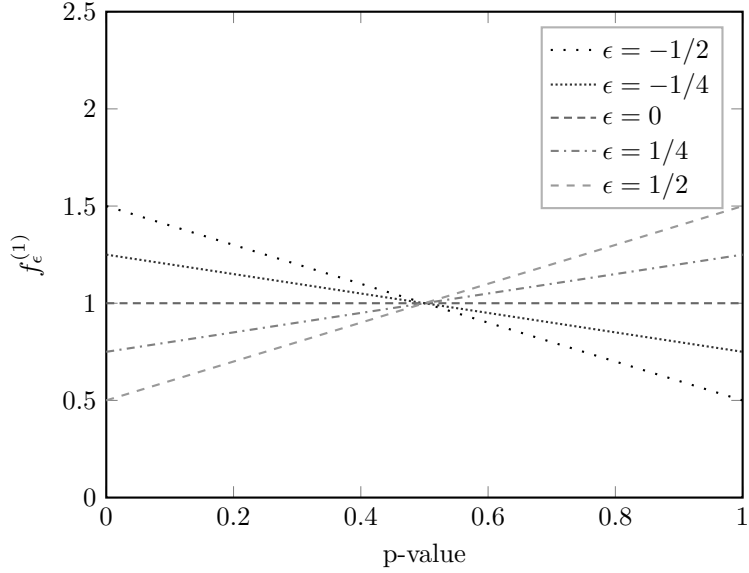

Any valid betting function $f$ must satisfy two constraints. It must integrate to 1, i.e. $\int_0^1f(p)dp=1$, and be non-negative, i.e. $f(p)\geq0$ for all $p\in[0,1]$. The betting function \eqref{eq:SimleLegendreJumper-bf} trivially satisfies the integral constraint as the shifted Legendre polynomial integrates to zero on the unit interval by orthogonality. For the second condition to hold, we must ensure that $\epsilon$ is chosen suitably. Let $M_k = \max_{p\in[0,1]}\widetilde{P}_k(p)$ and $m_k = \min_{p\in[0,1]}\widetilde{P}_k(p)$. By definition, $M_k\geq1$ because $\widetilde{P}_k(1) = 1$. For $f_{\epsilon}^{(k)}(p)\geq0$ to hold, we must have
\begin{equation}
	\epsilon\geq-\frac{1}{M_k} \quad \text{and}\quad \epsilon\geq - \frac{1}{m_k}.
\end{equation}
We will return to these constraints in Secton \ref{sec:state-space}, but for now, it suffices to state that the standard grid $\EE$ lies within the constraints for all $k$. The Simple Legendre Jumper martingale can be computed by Algorithm \ref{alg:SLJ}.
\begin{algorithm}[ht]
	\caption{Simple Legendre Jumper betting martingale}
	\label{alg:SLJ}
	\begin{algorithmic}
		\State \textbf{Parameters:} state space $\EE$, jumping rate $J\in(0,1]$, and Legendre degree $k\geq1$.
		\State \textbf{Input:} p-values $p_1,p_2,\dots$.
		\State \textbf{Output:} martingale $S_1,S_2,\dots$..
		\State $C_{\epsilon} \gets 1/|\EE|$ for all $\epsilon\in\EE$
		\State $C \gets 1$
		\For{$n = 1, 2, \dots$}
		\For{$\epsilon \in \EE$:
			$C_{\epsilon} \gets (1-J) C_{\epsilon} + (J/|\EE|)C$}
		\EndFor
		\For{$\epsilon \in \EE$:
		$C_{\epsilon} \gets C_{\epsilon}f_{\epsilon}^{(k)}(p_n)$}
		\EndFor
		\State $S_n \gets C := \sum_{\epsilon\in\EE}C_{\epsilon}$
		\EndFor
	\end{algorithmic}
\end{algorithm}
As noted by \citet{Bostrom2026}, the Simple Jumper (and thus the Simple Legendre Jumper of order $k=1$) has limited power against non-mean shifts. The reason for this is clear from Figure \ref{fig:betting-functions-k1}: $\epsilon<0$ corresponds to betting on small p-values, and $\epsilon>0$ corresponds to betting on large p-values. A symmetric redistribution of p-values---such as increased variance without a shift in mean---leaves both bets roughly balanced, yielding little net evidence.

\begin{figure}[ht]
	\centering
	\begin{tikzpicture}[
			declare function={
					P1(\p) = 2*\p - 1;
					P2(\p) = 6*\p^2 - 6*\p + 1;
					P3(\p) = 20*\p^3 - 30*\p^2 + 12*\p - 1;
					P4(\p) = 70*\p^4 - 140*\p^3 + 90*\p^2 - 20*\p + 1;
					f(\p,\eps) = 1 + \eps * P2(\p);
				}
		]
		\begin{axis}[
				width=10cm,
				height=8cm,
				xmin=0, xmax=1,
				ymin=0, ymax=2.5,
				xlabel={p-value},
				ylabel={$f_\epsilon^{(1)}$},
				legend pos=north east,
				legend style={draw=black!30},
				legend cell align={left},
				domain=0:1,
				samples=100,
				thick 
			]

			\addplot [black, loosely dotted] {f(x, -1/2)};
			\addlegendentry{$\epsilon = -1/2$}

			\addplot [black!80, densely dotted] {f(x, -1/4)};
			\addlegendentry{$\epsilon = -1/4$}

			\addplot [black!60, densely dashed] {f(x, 0)};
			\addlegendentry{$\epsilon = 0$}

			\addplot [black!50, dashdotted] {f(x, 1/4)};
			\addlegendentry{$\epsilon = 1/4$}

			\addplot [black!40, dashed] {f(x, 1/2)};
			\addlegendentry{$\epsilon = 1/2$}

		\end{axis}
	\end{tikzpicture}
	\caption{Betting functions of the Simple Legendre Jumper for $k=2$.}
	\label{fig:betting-functions-k2}
\end{figure}
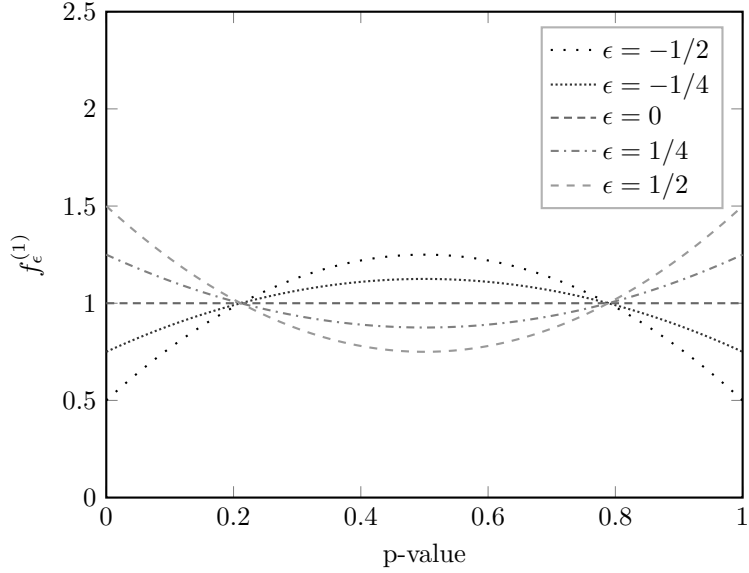

To enable the Simple Legendre Jumper to profit from a shift that is restricted purely to the variance, we can employ the shifted Legendre polynomial of degree 2. The resulting betting functions on the standard grid $\EE$ are illustrated in Figure \ref{fig:betting-functions-k2}. Here, $\epsilon<0$ corresponds to betting on p-values concentrated at $1/2$, whereas $\epsilon>1$ corresponds to betting on p-values concentrated at the boundaries. For higher orders, order 3 bets on changes in \emph{skewness}, order $4$ bets on changes in \emph{kurtosis}, and higher orders bet on increasingly complex multimodal asymmetric deviations from uniformity.

\begin{figure}[ht]
	\centering
	\includegraphics[width=\linewidth]{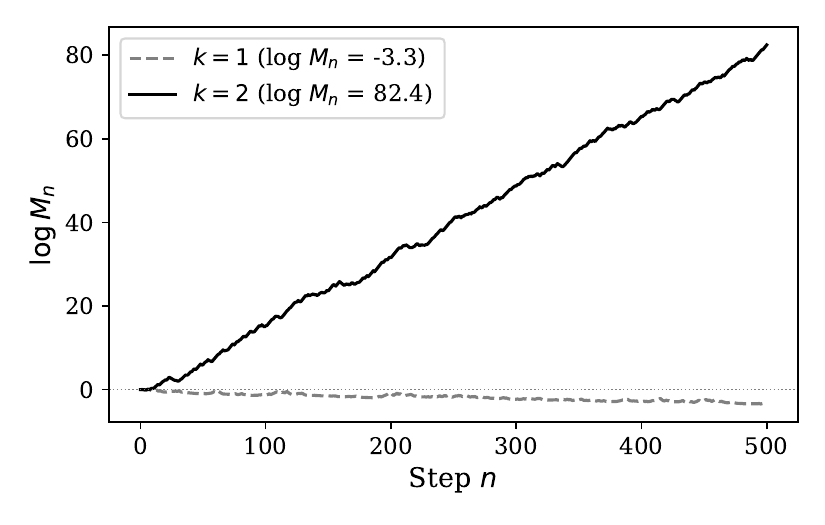}
	\caption{Trajectories of two Simple Legendre Jumper martingales of degrees $k=1,2$ on p-values drawn from the beta distribution $B(0.3,0.3)$. Degree 1 decreses while degree 2 grows exponentially.}
	\label{fig:SLJ-k1-vs-k2}
\end{figure}

Figure \ref{fig:SLJ-k1-vs-k2} illustrates the trajectories of two Simple Legendre Jumper martingales on p-values generated from the beta distribution $B(0.3,0.3)$, which preserves the mean $1/2$ of the uniform distribution but dramatically decreases the variance. Consequently, the SLJ of degree one (equivalent to the Simple Jumper) is unable to detect the deviation, but degree two grows exponentially.

\section{Product Legendre Jumper}\label{sec:plj}
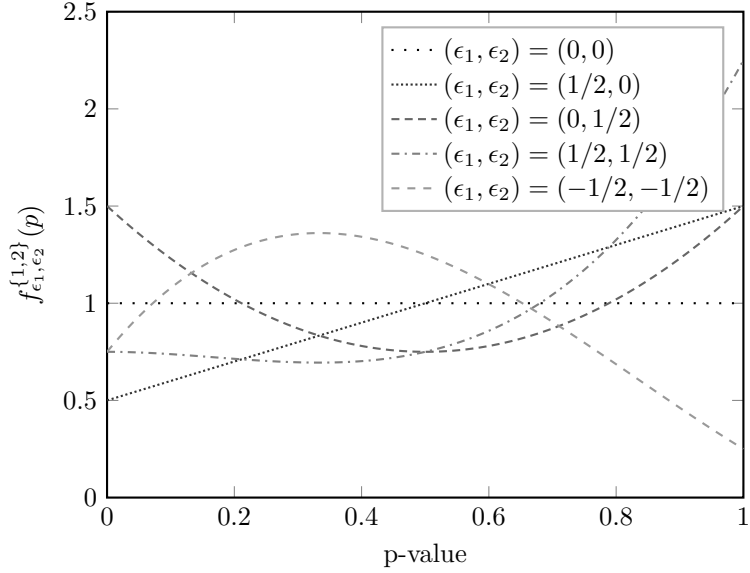
\begin{figure}[ht]
	\centering
	\begin{tikzpicture}[
			declare function={
					P1(\p) = 2*\p - 1;
					P2(\p) = 6*\p^2 - 6*\p + 1;
					F(\p,\eOne,\eTwo) = (1 + \eOne * P1(\p)) * (1 + \eTwo * P2(\p));
				}
		]
		\begin{axis}[
				width=10cm,
				height=8cm,
				xmin=0, xmax=1,
				ymin=0, ymax=2.5,
				xlabel={p-value},
				ylabel={$f_{\epsilon_1,\epsilon_2}^{\{1,2\}}(p)$},
				legend pos=north east,
				legend style={draw=black!30},
				legend cell align={left},
				domain=0:1,
				samples=100,
				thick 
			]

			\addplot [black, loosely dotted] {F(x, 0, 0)};
			\addlegendentry{$(\epsilon_1, \epsilon_2) = (0, 0)$}

			\addplot [black!80, densely dotted] {F(x, 1/2, 0)};
			\addlegendentry{$(\epsilon_1, \epsilon_2) = (1/2, 0)$}

			\addplot [black!60, densely dashed] {F(x, 0, 1/2)};
			\addlegendentry{$(\epsilon_1, \epsilon_2) = (0, 1/2)$}

			\addplot [black!50, dashdotted] {F(x, 1/2, 1/2)};
			\addlegendentry{$(\epsilon_1, \epsilon_2) = (1/2, 1/2)$}

			\addplot [black!40, dashed] {F(x, -1/2, -1/2)};
			\addlegendentry{$(\epsilon_1, \epsilon_2) = (-1/2, -1/2)$}

		\end{axis}
	\end{tikzpicture}
	\caption{Some betting functions of the Product Legendre Jumper for $k=1,2$.}
	\label{fig:product-betting-functions}
\end{figure}

The Simple Legendre Jumper lets us pick up on deviations from uniformity that do not result in a shifted mean. However, the construction does not make full use of the orthogonal system of polynomials. To do that, we need to modify the betting function further, and consider products of the betting functions \eqref{eq:SimleLegendreJumper-bf}. As a first step, we consider combining two Legendre polynomials of different degree, say $k=1,2$. The \emph{Product Legendre Jumper} (PLJ) uses the betting function
\begin{equation}\label{eq:product-betting-function-k12}
	f_{\epsilon_1,\epsilon_2}^{\{1,2\}}(p) = (1 + \epsilon\widetilde{P}_1(p))(1 + \epsilon\widetilde{P}_2(p)), \quad p\in[0,1]
\end{equation}
where $(\epsilon_1, \epsilon_2)\in\EE := \EE_1\times\EE_2$, and $\EE_1=\EE_2=\{-1/2,-1/4,0,1/4,1/2\}$. Expanding the product in \eqref{eq:product-betting-function-k12}, we see that
\begin{equation}
	f_{\epsilon_1,\epsilon_2}^{\{1,2\}}(p) =
	1 + \epsilon_1\widetilde{P}_1(p) + \epsilon_2\widetilde{P}_2(p) + \epsilon_1\epsilon_2\widetilde{P}_1(p)\widetilde{P}_2(p),
\end{equation}
which ensures that $\int_0^1f_{\epsilon_1,\epsilon_2}^{\{1,2\}}(p)dp=1$; the shifted Legendre polynomials are orthogonal, so the last term vanishes in the integral, and $\int_0^1\widetilde{P}_k(p)dp=0$ for $k\geq1$. Therefore, \eqref{eq:product-betting-function-k12} is a valid betting function if we can ensure that it is non-negative. We will return to this in Section \ref{sec:state-space}, but for now it suffices to note that if both factors in \eqref{eq:product-betting-function-k12} are non-negative, so is the product. Therefore, the Cartesian product of the standard $\epsilon$ grid is safe to use. Some of the betting functions determined by $\EE$ are illustrated in Figure \ref{fig:product-betting-functions}.

The Product Legendre Jumper depends additionally on the single jumping rate parameter $J\in(0,1]$.
Writing $\boldsymbol{\epsilon} = (\epsilon_1, \epsilon_2)\in\EE$, for any probability measure $\mu$ on $\EE^{\infty}$, the function
\begin{equation}\label{eq:betting-martingale-two-factors}
	F(p_1,\dots,p_n) := \int\prod_{i=1}^n
	f_{\epsilon_1,\epsilon_2}^{\{1,2\}}(p_i)\mu(d({\boldsymbol{\epsilon}}_1,{\boldsymbol{\epsilon}}_2, \dots))
\end{equation}
is a CTM. For the Product Legendre Jumper, the measure is the probability distribution of the Markov chain with state space $\EE$ which has initial state $\boldsymbol{\epsilon}$ chosen from the uniform distribution on $\EE$, maintains the same state with probability $1-J$, and chooses a new state from the uniform probability distribution on $\EE$ with probability $J$.

Of course, nothing stops us from adding shifted Legendre polynomials of higher order to the product. Let $K$ be a set of positive integers, representing the degrees included in the product, and $\boldsymbol{\epsilon} = (\epsilon_k)_{k\in K}$. The general Product Legendre betting function is then
\begin{equation}\label{eq:product-betting-function}
	f_{\boldsymbol{\epsilon}}^{K}(p) := \frac{\prod_{k\in K}(1+\epsilon_k\widetilde{P}_k(p))}{Z(\boldsymbol{\epsilon})}, \quad p\in[0,1],
\end{equation}
where
\begin{equation}
	Z(\boldsymbol{\epsilon}) = \int_0^1 \prod_{k\in K}(1+\epsilon_k\widetilde{P}_k(p)).
\end{equation}

To understand why this normalisation constant $Z(\boldsymbol{\epsilon})$ is necessary for $|K| \geq 3$, consider the expansion of the product for $K = \{1, 2, 3\}$. While the integrals of the individual polynomials and the pairwise cross-terms vanish due to orthogonality, the product of three polynomials does not generally integrate to zero. Specifically, expanding the product and integrating yields
\begin{equation}
	Z(\boldsymbol{\epsilon}) = 1 + \epsilon_1 \epsilon_2 \epsilon_3 \int_0^1 \widetilde{P}_1(p)\widetilde{P}_2(p)\widetilde{P}_3(p)dp.
\end{equation}
We can evaluate this integral exactly by shifting back to the standard symmetric interval $[-1, 1]$ via the substitution $x = 2p - 1$:
\begin{equation}
    \begin{aligned}
	   \frac{1}{2}& \int_{-1}^1 P_1(x)P_2(x)P_3(x)dx \\&= \frac{1}{2} \int_{-1}^1 x \left(\frac{3x^2 - 1}{2}\right) \left(\frac{5x^3 - 3x}{2}\right) dx.
    \end{aligned}
\end{equation}
Expanding the factors yields $\frac{1}{8} \int_{-1}^1 (15x^6 - 14x^4 + 3x^2) dx = \frac{3}{35}$. Thus, the normalisation constant for this specific product is exactly $Z(\boldsymbol{\epsilon}) = 1 + \frac{3}{35}\epsilon_1\epsilon_2\epsilon_3$.

In general, expanding the product in \eqref{eq:product-betting-function} and integrating term by term yields
\begin{equation}\label{eq:Z-expansion}
	Z(\boldsymbol{\epsilon}) = 1 + \sum_{\substack{S \subseteq K \\ |S| \geq 3}} G_S \prod_{k \in S} \epsilon_k,
\end{equation}
where
\begin{equation}\label{eq:gaunt-coeff}
	G_S := \int_0^1 \prod_{k \in S} \widetilde{P}_k(p)\,dp.
\end{equation}
Terms of just one polynomial one vanish since $\int_0^1 \widetilde{P}_k(p)\,dp = 0$, and terms containing two polynomials vanish by orthogonality, so only subsets of size $|S| \geq 3$ contribute. The coefficients $G_S$ are rational numbers computable in closed form, since the product of polynomials is once again a polynomial. For three polynomials, this is historically known as Gaunt's formula, which evaluates to the square of the Wigner 3-j symbol \citep{edmonds1985angular}. Table \ref{tab:gaunt} lists all non-zero coefficients for $K \subseteq \{1,2,3,4\}$; note that $G_S = 0$ whenever $\sum_{k \in S} k$ is odd. To see this, observe that after the substitution $x=2p-1$, the integral becomes a product of standard Legendre polynomials of odd total degree over $[-1,1]$. Since a product of Legendre polynomials of odd combined degree is an odd function (see \eqref{eq:parity}), and odd functions integrate to zero over the symmetric interval $[-1,1]$, we have $G_S = 0$.
\begin{table}[ht]
	\centering
	\begin{tabular}{c|c}
		$S$           & $G_S$      \\
		\hline
		$\{1,2,3\}$   & $3/35$     \\
		$\{1,3,4\}$   & $4/63$     \\
		$\{1,2,3,4\}$ & $142/3465$
	\end{tabular}
	\caption{Non-zero coefficients $G_S$ for subsets $S \subseteq \{1,2,3,4\}$.}
	\label{tab:gaunt}
\end{table}

Since the grid $\EE$ is finite, all $|\EE|$ values of $Z(\boldsymbol{\epsilon})$ can be pre-computed at initialisation and stored as a lookup table, making per-step evaluation $O(1)$. However, as the size of $K$ increases, the number of non-vanishing higher-order cross terms grows rapidly. Although closed-form solutions exist for all resulting integrals, tracking the combinatorial expressions becomes algebraically cumbersome. Therefore, to maintain clarity in our exposition while sufficiently demonstrating the mechanics of the joint deformations, we limit our further explicit derivations to combinations within $K \subseteq \{1,2,3\}$.

The state space of the Markov chain for $|K|=k$ is $\EE = \prod_{k\in K}\EE_k$, where $\EE_k$ is our standard grid $\{-1/2,-1/4,0,1/4,1/2\}$. The Product Legendre Jumper martingale can be computed using Algorithm \ref{alg:PLJ}.
\begin{algorithm}[ht]
	\caption{Product Legendre Jumper betting martingale}
	\label{alg:PLJ}
	\begin{algorithmic}
		\State \textbf{Parameters:} set of degrees $K$, state space $\EE=\prod_{k\in K}E_k$, and jumping rate $J\in(0,1]$. \Comment{Note: $\boldsymbol{\epsilon} = (\epsilon_k)_{k\in K}$ is a vector.}
		\State \textbf{Input:} p-values $p_1,p_2,\dots$.
		\State \textbf{Output:} martingale $S_1,S_2,\dots$..
		\State $C_{\boldsymbol{\epsilon}} \gets 1/|\EE|$ for all $\boldsymbol{\epsilon}\in\EE$
		\State $C \gets 1$
		\For{$n = 1, 2, \dots$}
		\For{$\boldsymbol{\epsilon} \in \EE$:
			$C_{\boldsymbol{\epsilon}} \gets (1-J) C_{\boldsymbol{\epsilon}} + (J/|\EE|)C$}
		\EndFor
		\For{$\boldsymbol{\epsilon} \in \EE$:
		$C_{\boldsymbol{\epsilon}} \gets C_{\boldsymbol{\epsilon}}f_{\boldsymbol{\epsilon}}^{K}(p_n)$}
		\EndFor
		\State $S_n \gets C := \sum_{\boldsymbol{\epsilon}\in\EE}C_{\boldsymbol{\epsilon}}$
		\EndFor
	\end{algorithmic}
\end{algorithm}

\begin{figure}[ht]
	\centering
	\includegraphics[width=\linewidth]{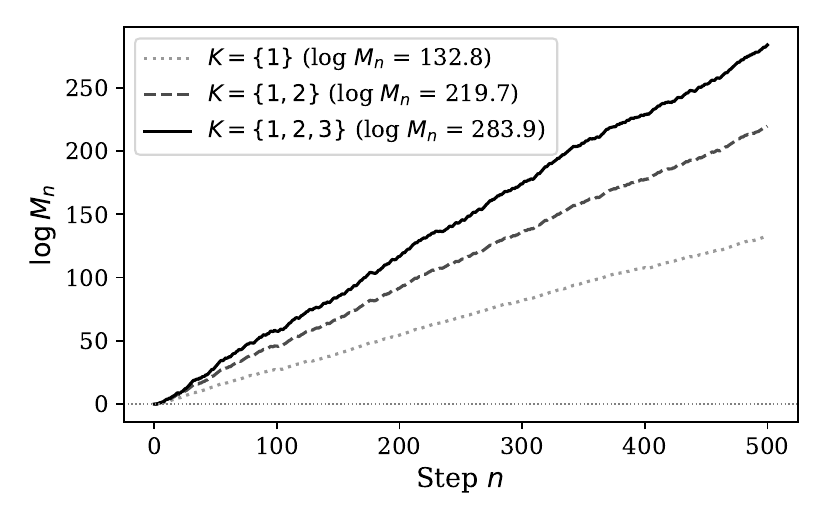}
	\caption{Trajectories of Product Legendre Martingales with degrees $K=\{1\},\{1,2\},\{1,2,3\}$ on p-values generated from the beta distribution $B(0.3,1.5)$.}
	\label{fig:PLJ-k1-vs-k12-vs-k123}
\end{figure}

Figure \ref{fig:PLJ-k1-vs-k12-vs-k123} shows the trajectories of the Product Legendre Martingales with degrees $K=\{1\},\{1,2\},\{1,2,3\}$ on p-values generated from the beta distribution $B(0.3,1.5)$ which differs from the uniform distribution in terms of mean, variance, and skewness. In this case, all martingales grow, but adding more degrees results in quicker growth.

Although Algorithm \ref{alg:PLJ} captures complex joint deviations from uniformity, it reveals a severe computational and statistical bottleneck embedded within the standard Markov transition mechanism. We refer to this as the \emph{jumping tax}.
To see precisely where this tax is paid, consider the wealth redistribution step in the first inner loop of Algorithm \ref{alg:PLJ}:
\begin{equation}\label{eq:jump-tax}
	C_{\boldsymbol{\epsilon}} \gets (1-J) C_{\boldsymbol{\epsilon}} + \frac{J}{|\EE|}C.
\end{equation}
When the environment undergoes a distribution shift, the martingale relies on the jumping mechanism to move capital to the newly optimal state. However, the available jumping capital $J \cdot C$ is distributed uniformly across the entire state space $\EE$.

If we employ a grid of resolution $g = |\EE_k|$ (e.g. $g=5$) across all degrees $k \in K$, the size of the joint state space scales exponentially: $|\EE| = g^K$. Thus, the newly optimal joint expert $\boldsymbol{\epsilon}^*$ receives only the fraction $J / g^K$ of the global capital. For instance, if $K=\{1,2,3,4\}$ and $g=5$, the target state receives a minuscule $J/625$ of the wealth. The martingale dilutes its capital by uniformly probing suboptimal orthogonal combinations, thereby delaying the detection of non-stationarity.

To retain the geometric expressiveness of higher-degree Legendre polynomials while avoiding the exponential jump tax, we decouple the adaptation of each polynomial degree. We term this the Variational Legendre Jumper.

\subsection{Variational Legendre Jumper}\label{sec:vlj}
Inspired by the mean-field approximation in variational inference \citep{wainwright2008graphical, blei2017variational}, we decompose the joint adaptation problem into $|K|$ independent sub-problems. Rather than maintaining a single Markov chain over the exponentially large product space $\EE$, we run $|K|$ independent instances of Algorithm~\ref{alg:SLJ}---one for each polynomial degree $k \in K$---each adapting its own parameter $\bar{\epsilon}_k$ on the marginal state space $\EE_k$.
At each step, the sub-jumpers provide a vector of \emph{consensus parameters} $\bar{\boldsymbol{\epsilon}} = (\bar{\epsilon}_k)_{k \in K}$, computed as the wealth-weighted mean of each marginal state:
\begin{equation}\label{eq:consensus}
	\bar{\epsilon}_k = \sum_{\epsilon \in \EE_k} \epsilon \frac{C_{k,\epsilon}}{C_k}.
\end{equation}
The \emph{Variational Legendre Jumper} then bets with the normalised product betting function evaluated at these consensus parameters
\begin{equation}\label{eq:vlj-bet}
	S_n = S_{n-1} \cdot f_{\bar{\boldsymbol{\epsilon}}}^K(p_n) = S_{n-1} \cdot \frac{\prod_{k \in K}(1 + \bar{\epsilon}_k \tilde{P}_k(p_n))}{Z(\bar{\boldsymbol{\epsilon}})}.
\end{equation}
Because $\bar{\boldsymbol{\epsilon}}$ is determined entirely by the sub-jumper states before observing $p_n$ (and hence is predictable), the process $S_n$ is a valid test martingale.

Each sub-jumper updates its internal state using its own marginal betting function $f_{\epsilon}^{(k)}(p) = 1 + \epsilon\tilde{P}_k(p)$, exactly as in Algorithm~\ref{alg:SLJ}. The sub-jumpers are completely decoupled from each other and from the global martingale; they serve purely as an adaptive mechanism for producing consensus parameters.
The computational cost is $O(|K| \cdot g)$ per step, regardless of $|K|$. This eliminates the exponential scaling of Algorithm~\ref{alg:PLJ}, making it feasible to include arbitrarily many polynomial degrees. The complete procedure is presented in Algorithm~\ref{alg:VLJ}.

\begin{algorithm}[ht]
	\caption{Variational Legendre Jumper betting martingale}
	\label{alg:VLJ}
	\begin{algorithmic}
		\State \textbf{Parameters:} set of degrees $K$, state spaces $\EE_k$, and jumping rate $J\in(0,1]$.
		\State \textbf{Input:} p-values $p_1,p_2,\dots$.
		\State \textbf{Output:} martingale $S_1,S_2,\dots$.
		\State $C_{k, \epsilon} \gets 1/|\EE_k|$ for all $k \in K, \epsilon \in \EE_k$
		\State $C_k \gets 1$ for all $k \in K$
		\State $S_0 \gets 1$
		\For{$n = 1, 2, \dots$}
		\For{$k \in K, \epsilon \in \EE_k$:
			$C_{k, \epsilon} \gets (1-J) C_{k, \epsilon} + (J/|\EE_k|)C_k$} \Comment{Jump (per sub-jumper)}
		\EndFor
		\For{$k \in K$:
			$\bar{\epsilon}_{k} \gets \sum_{\epsilon \in \EE_k} \epsilon(C_{k, \epsilon}/C_k)$} \Comment{Consensus parameters}
		\EndFor
		\State $S_n \gets S_{n-1} \cdot f_{\bar{\boldsymbol{\epsilon}}}^K(p_n)$ \Comment{Global bet (normalised by $Z$)}
		\For{$k \in K, \epsilon \in \EE_k$:
			$C_{k, \epsilon} \gets C_{k, \epsilon} \cdot f_{\epsilon}^{(k)}(p_n)$} \Comment{Update sub-jumpers}
		\EndFor
		\For{$k \in K$:
			$C_k \gets \sum_{\epsilon \in \EE_k} C_{k, \epsilon}$}
		\EndFor
		\EndFor
	\end{algorithmic}
\end{algorithm}

\begin{remark}
	For $|K| \leq 2$, the normalisation constant satisfies $Z(\bar{\boldsymbol{\epsilon}}) = 1$ by Legendre orthogonality. In this case, the global bet simplifies to $S_n = S_{n-1} \cdot \prod_k (1 + \bar{\epsilon}_k \tilde{P}_k(p_n))$, and since each sub-jumper's growth factor is precisely $C_k^{(n)}/C_k^{(n-1)} = 1 + \bar{\epsilon}_k \tilde{P}_k(p_n)$, the martingale reduces to $S_n = \prod_{k \in K} C_k$. That is, the global martingale is exactly the product of the individual sub-jumper martingales---each an independent instance of Algorithm~\ref{alg:SLJ}.
\end{remark}

\begin{figure}[ht]
	\centering
	\includegraphics[width=\linewidth]{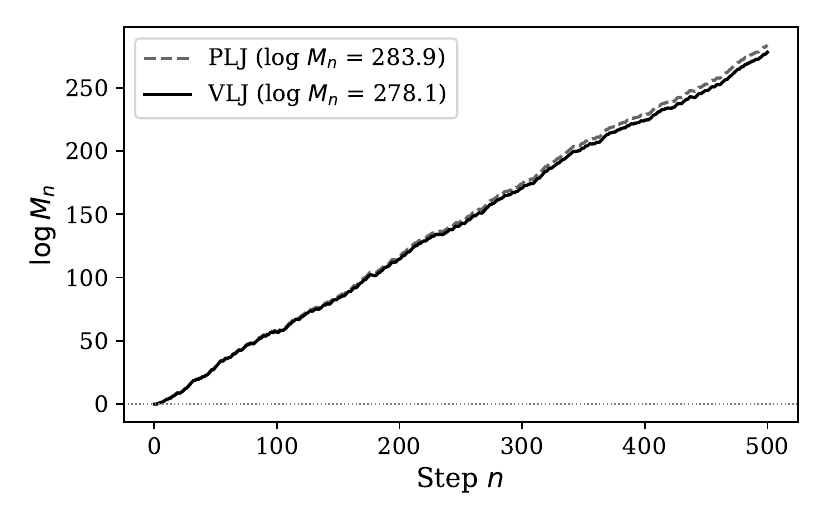}
	\caption{Trajectories of the Product Legendre Jumper and the Variational Legendre Jumper, both with $K=\{1,2,3\}$, processing p-values generated from the beta distribution $B(0.3,1.5)$.}
	\label{fig:plj-vs-vlj-d123}
\end{figure}

Figure \ref{fig:plj-vs-vlj-d123} illustrates the trajectories of the Product Legendre Jumper and the Variational Legendre Jumper, both with $K=\{1,2,3\}$, processing p-values generated from the beta distribution $B(0.3,1.5)$. The trajectories are almost indistinguishable until about $n=200$, after which PLJ takes the lead and reaches a higher final value. Overall, PLJ and VLJ perform remarkably similar in this example.

\begin{figure}[ht]
	\centering
	\includegraphics[width=\linewidth]{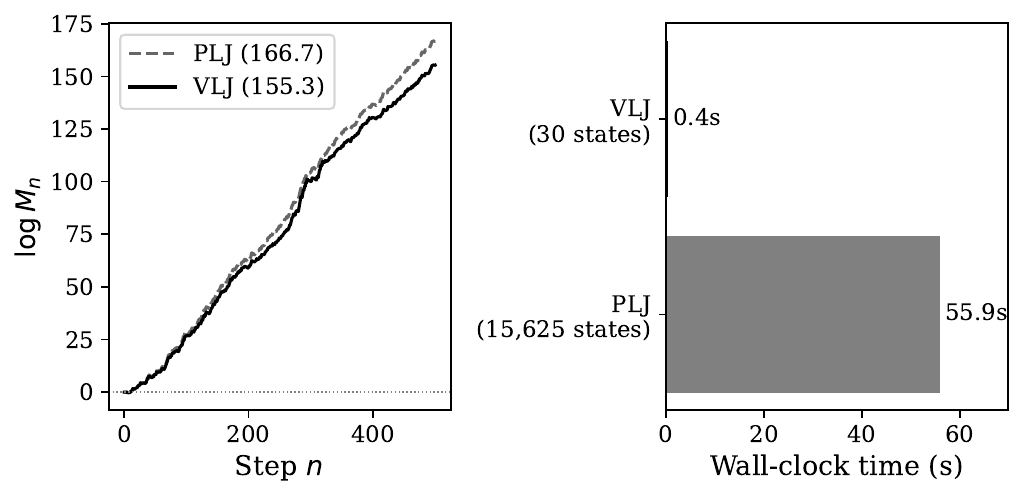}
	\caption{Trajectories of the Product Legendre Jumper and the Variational Legendre Jumper, both with $K=\{1,2,3,4,5,6\}$, processing p-values generated from the mixture $0.4 \cdot B(0.2, 0.8) + 0.6 \cdot B(2.0, 0.5)$ (left) and the wall-clock time elapsed to compute the trajectories (right). Timings measured on a 12th Gen Intel Core i7-1255U (40GiB RAM).}
	\label{fig:plj-vs-vlj-d123456-time}
\end{figure}

Figure \ref{fig:plj-vs-vlj-d123456-time} shows the difference in the per-step computational effort involved with the Product Legendre Jumper ($O(|K|g^{|K|})$) and Variational Legendre Jumper ($O(|K|g)$). The left-hand side plot shows the trajectories of the PLJ and VLJ, both with $K=\{1,2,3,4,5,6\}$, and the right-hand side plot presents the wall-clock time per trajectory.

\section{The Composite Legendre Jumper}\label{sec:composite}
In practice, the optimal jumping rate $J$ depends on the timescale of the distributional shift---abrupt changes favour larger $J$, while gradual drift favours smaller $J$---and this timescale is rarely known in advance. Following \citet{alrw2}, we define the \emph{Composite Legendre Jumper} as the arithmetic mean of the base martingale over the rates
\[
	J \in \{10^{-4}, 10^{-3}, 10^{-2}, 10^{-1}, 1\}.
\]
The construction applies to any base martingale---SLJ, PLJ, or VLJ---and the resulting average is itself a valid martingale by convexity (a convex combination of martingales is a martingale).

We follow \citet{alrw2} in constructing the grid of jumping rates. Other choices are valid and could potentially be beneficial, although Vovk et al. remarked that the dependence of the Simple Jumper on the choice of jumping rate is not heavy. The inclusion of $J = 1$ is important because $\EE$ is symmetric around zero; the sub-jumper at $J = 1$ resets to uniform weights every step, producing bets of exactly one and a martingale value that is identically equal to one. This guarantees that the composite martingale never drops below $1/|J| = 1/5$, providing an absolute wealth floor that is absent in any fixed-$J$ variant. Therefore, the Composite Legendre Jumper is the recommended default when the shift timescale is unknown.

\section{State space restrictions and the standard grid}\label{sec:state-space}
As established, any valid betting function $f$ must satisfy two constraints: it must integrate to 1, i.e., $\int_0^1 f(p)dp = 1$, and it must be non-negative, $f(p) \geq 0$, for all $p \in [0,1]$. The integral constraint is automatically satisfied by all our Legendre constructions, as we have already noted.

The non-negativity constraint, however, requires careful parameter bounding, particularly for the product betting functions. Although a product can technically be non-negative if multiple factors are simultaneously negative, tracking these joint boundaries across higher dimensions is analytically cumbersome. A much simpler, sufficient condition is to restrict our attention to ensuring that every individual factor is strictly non-negative. If $1 + \epsilon_k \widetilde{P}_k(p) \geq 0$ for all $k$, their product is guaranteed to be non-negative in all cases. Therefore, we limit our consideration to bounding the individual parameters $\epsilon$ for the Simple Legendre Jumper betting function \eqref{eq:SimleLegendreJumper-bf}, from which the validity of the rest follows.

To keep $1 + \epsilon\widetilde{P}_k(p) \geq 0$, we require that $\epsilon \leq -1/\min_p \widetilde{P}_k(p)$ when $\epsilon > 0$, and $\epsilon \geq -1/\max_p \widetilde{P}_k(p)$ when $\epsilon < 0$.
A fundamental property of the standard Legendre polynomials is that they are strictly bounded such that $|P_k(x)| \leq 1$ for all $x \in [-1,1]$ and all degrees $k$ \citep{szeg1939orthogonal}. Consequently, for the shifted polynomials defined on $[0,1]$, the global maximum is always exactly $\max_p \widetilde{P}_k(p) = \widetilde{P}_k(1) = 1$ (recall that $\widetilde{P}_k(1)=1$ by definition). The global minimum is bounded by $\min_p \widetilde{P}_k(p) \geq -1$ (where equality holds exactly for odd $k$ at $p=0$).
Applying these extrema to our inequalities yields a simple, universal bound: the betting function $f_{\epsilon}^{(k)}(p) = 1 + \epsilon\widetilde{P}_k(p)$ is strictly non-negative for all $p \in [0,1]$ and all degrees $k$ if and only if
\begin{equation}
	\epsilon \in [-1, 1].
\end{equation}
Because our chosen standard grid $\EE = \{-1/2, -1/4, 0, 1/4, 1/2\}$ is strictly contained within this interval, it is guaranteed to be a valid parameter space for a Simple Legendre Jumper of any arbitrary degree $k$. Furthermore, by our sufficient condition above, the Cartesian product of this grid safely guarantees non-negativity for the Product and Variational Legendre Jumpers as well.

\section{Empirical results}\label{sec:empirical}
We evaluated the Legendre Jumper martingales on a real-world classification task using the Wine Quality dataset \citep{cortez2009wine}, available from the UCI Machine Learning Repository (ID 186). The dataset contains 6497 samples (1599 red wines and 4898 white wines) described by 11 physicochemical features (e.g., acidity, residual sugar, and alcohol content). Each wine is assigned a quality score in $\{3, 4, 5, 6, 7, 8, 9\}$, which we treat as a 7-class label space. All features were standardised to zero mean and unit variance.

We employed a conformal nearest-neighbor classifier with $k=1$ \citep[Section 3.2.1]{alrw2} in the full online conformal prediction protocol. The nonconformity measure computes, for each example in the bag, the ratio of the Euclidean distance to its nearest same-class neighbour to the distance to its nearest different-class neighbour. An object is thus considered nonconforming if it is close to an object labelled differently and far from any object sharing its label. An initial training set of $1599$ samples was used, after which the remaining 4898 observations were processed sequentially. We set the jumping rate $J = 0.01$ and used the standard grid $\EE = \{-1/2, -1/4, 0, 1/4, 1/2\}$ throughout.

To probe different types of distributional shift, we consider four orderings of the data:
\begin{enumerate}
	\item \textbf{Shuffled:} a uniformly permutation of the whole dataset, enforcing exchangeability (p-values should be uniform).
	\item \textbf{Red $\to$ White:} all red wines first (randomly shuffled within the group), then all white wines (also randomly shuffled), creating an abrupt change-point at the start of the test stream.
	\item \textbf{White $\to$ Red:} the reverse arrangement (each group again internally shuffled), with a change-point occurring approximately halfway through the test stream.
	\item \textbf{Original order:} the dataset in its original UCI row order, in which all red wines appear first followed by all white wines, preserving within-group structure.
\end{enumerate}

We compare seven martingale configurations: the Simple Legendre Jumper at orders $k = 1, 2, 3$ individually; the Product Legendre Jumper with $K = \{1, 2\}$ and $K = \{1, 2, 3\}$; and the Variational Legendre Jumper with the same two sets of degrees.

\begin{figure}[ht]
	\centering
	\includegraphics[width=\linewidth]{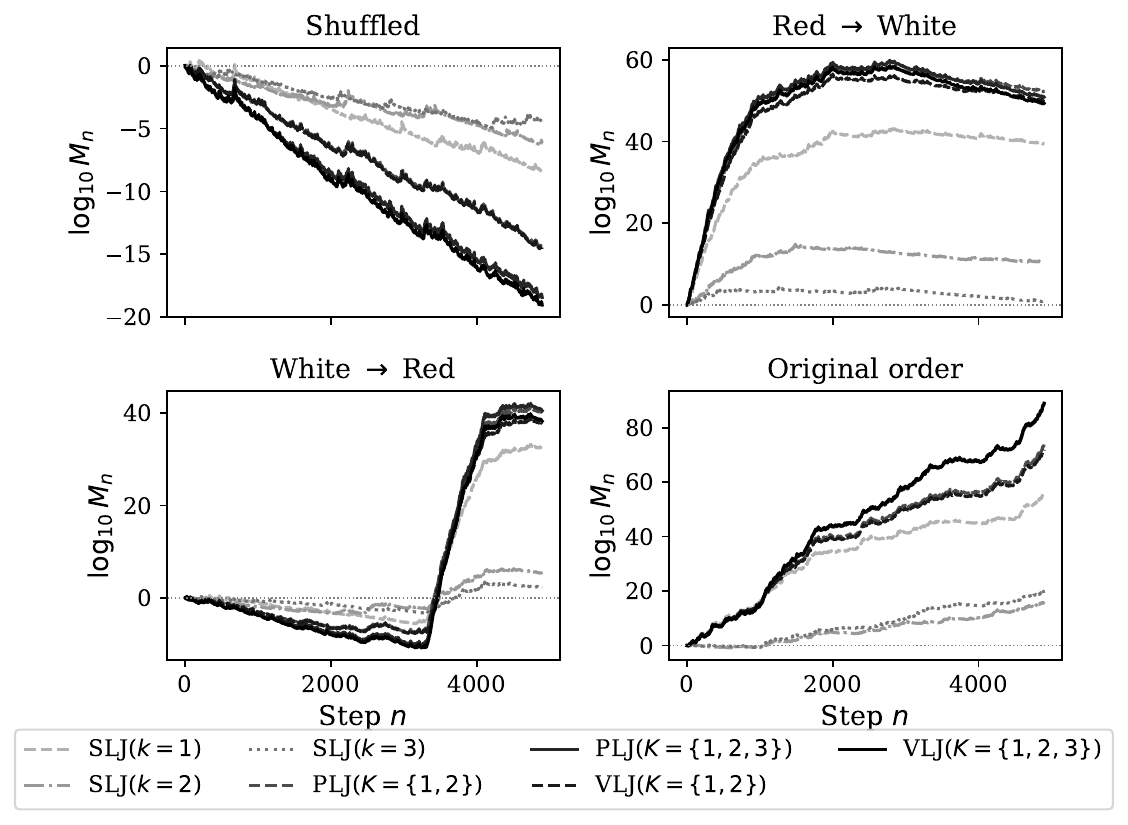}
	\caption{Log-martingale trajectories ($\log_{10} M_n$) for seven Legendre Jumper configurations on the Wine Quality dataset under four orderings. Under exchangeability (top-left), all martingales remain near zero. Under distributional shift, PLJ and VLJ consistently dominate SLJ($k{=}1$), while SLJ($k{=}2$) and SLJ($k{=}3$) alone accumulate comaratively little evidence.}
	\label{fig:wine-nn-experiment}
\end{figure}

\begin{table}[ht]
	\centering
	\begin{tabular}{l rrr rr rr}
		\hline
		                & \multicolumn{3}{c}{SLJ} & \multicolumn{2}{c}{PLJ} & \multicolumn{2}{c}{VLJ}                                                     \\
		Ordering        & $k{=}1$                 & $k{=}2$                 & $k{=}3$                 & $\{1,2\}$ & $\{1,2,3\}$ & $\{1,2\}$ & $\{1,2,3\}$ \\
		\hline
		Shuffled        & $-8.4$                  & $-6.1$                  & $-4.4$                  & $-14.4$   & $-18.4$     & $-14.6$   & $-19.1$     \\
		Red $\to$ White & $39.4$                  & $10.4$                  & $1.1$                   & $52.1$    & $50.9$      & $49.8$    & $49.3$      \\
		White $\to$ Red & $32.4$                  & $5.3$                   & $2.2$                   & $40.0$    & $40.5$      & $37.7$    & $38.3$      \\
		Original        & $56.2$                  & $15.8$                  & $19.9$                  & $73.8$    & $89.1$      & $72.0$    & $88.9$      \\
		\hline
	\end{tabular}
	\caption{Final $\log_{10} M_n$ after all 4898 test observations. Higher values indicate stronger evidence against exchangeability.}
	\label{tab:wine-nn-final}
\end{table}

The results are shown in Figure~\ref{fig:wine-nn-experiment} and Table~\ref{tab:wine-nn-final}. Under shuffled ordering, all martingales remain close to zero and decline slightly, confirming that they do not raise false alarms under exchangeability. Notably, the higher-order methods do not lose substantially more capital than SLJ($k{=}1$) despite operating over a larger state space, validating that the jumping tax is manageable.

Under the three non-exchangeable orderings, a clear hierarchy emerges. The Product and Variational Legendre Jumpers with $K = \{1,2,3\}$ consistently achieve the highest final martingale values, outperforming SLJ($k{=}1$) by 30\%--60\% in $\log_{10} M_n$. This demonstrates the advantage of simultaneously betting on multiple polynomial degrees: the combined methods capture more of the full distributional change, including shifts in both location and higher-order moments. In contrast, SLJ($k{=}2$) and SLJ($k{=}3$) alone perform poorly; betting exclusively on variance or skewness, they miss the dominant location shift signal and accumulate orders of magnitude less evidence.

Finally, the Variational Legendre Jumper matches the Product Legendre Jumper almost exactly across all orderings, confirming that the mean-field factorisation introduced in the variational approximation incurs a negligible loss in practice. This is particularly encouraging given the exponential reduction in computational cost: while PLJ($K{=}\{1,2,3\}$) maintains $|\EE|^3 = 125$ states, VLJ($K{=}\{1,2,3\}$) requires only $3|\EE| = 15$.

\begin{figure}[ht]
	\centering
	\includegraphics[width=\linewidth]{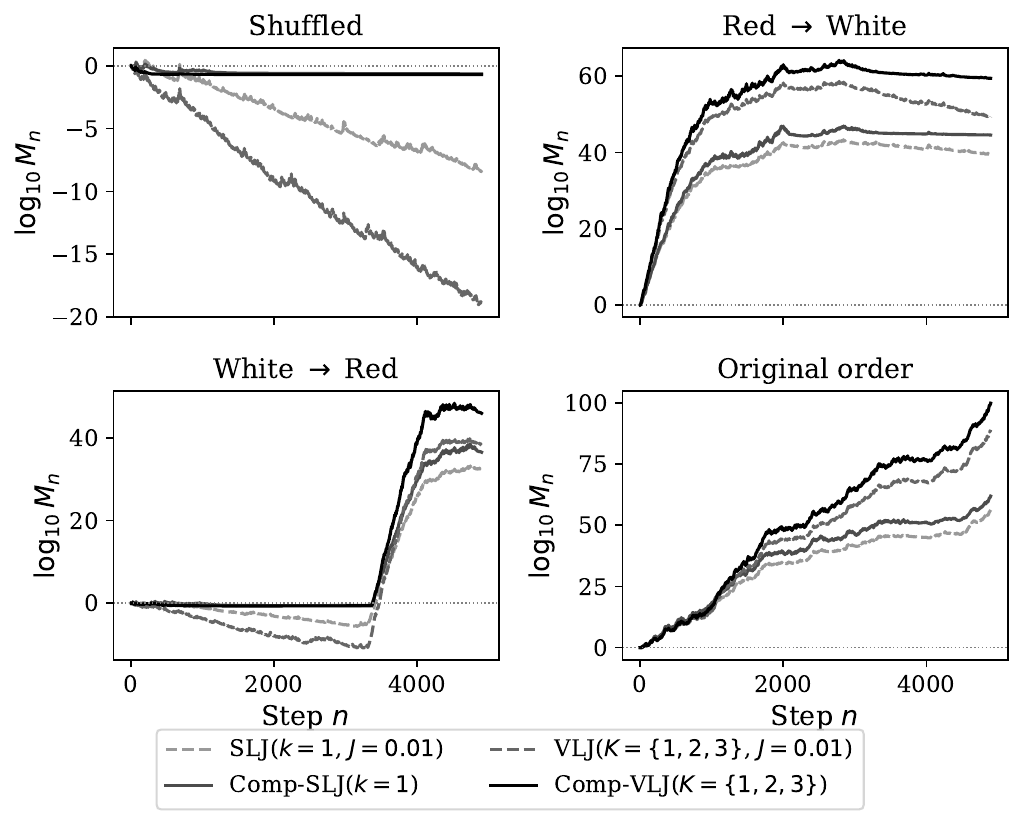}
	\caption{Fixed-$J$ versus Composite Legendre Jumpers on the Wine Quality dataset. Solid lines denote composite variants (averaging over five jumping rates); dashed lines denote fixed $J = 0.01$. The composite provides a wealth floor under exchangeability (top-left) and substantially outperforms the fixed rate under distributional shift.}
	\label{fig:wine-nn-composite}
\end{figure}

We additionally evaluate the Composite Legendre Jumper introduced in Section~\ref{sec:composite}. Figure~\ref{fig:wine-nn-composite} compares two pairs: SLJ($k{=}1$) at $J = 0.01$ versus its composite, and VLJ($K{=}\{1,2,3\}$) at $J = 0.01$ versus its composite.

Under the shuffled ordering, the composite variants remain near zero, whereas the fixed-$J$ martingales drift to $\log_{10} M_n \approx -8$ (SLJ) and $-20$ (VLJ). The $J = 1$ sub-jumper anchors the composite near its floor, preventing the slow accumulation of losses that affect the fixed-rate versions when there is no signal to exploit.

Under a distributional shift, the composite decisively dominates. On the original ordering, Comp-VLJ($K{=}\{1,2,3\}$) reaches $\log_{10} M_n \approx 100$, surpassing the fixed-$J$ VLJ by more than ten orders of magnitude, far exceeding the $\log_{10}(5) \approx 0.7$ mixing penalty. This indicates that the optimal jumping rate for these abrupt shifts is considerably faster than $0.01$; the composite automatically allocates most of its wealth to faster sub-jumpers without requiring the user to specify this in advance. Therefore, the Composite Legendre Jumper is the recommended default when the shift timescale is unknown.

So far, we have investigated the growth and maximum values attained by various Legendre Jumpers. We now address the practical deployment question of ML models in production: when should a model be retrained? This application uses a computationally efficient data-splitting approach rather than the fully sequential protocol above, but retains the same theoretical guarantees.

\subsection{When do we retrain a model?}\label{subsec:retrain-or-not}
The Wine Quality dataset was studied in \citet{vovk2021retrain}, where the authors developed a method for detecting distribution shifts and determining when a prediction algorithm requires retraining using the Simple Jumper martingale, which is equivalent to the Simple Legendre martingale of degree one. Vovk et al. investigated when a \emph{pre-trained model} should be retrained to avoid degrading its predictive performance. Therefore, they used \emph{inductive conformal transducers} \citep{papadopoulos2002inductive,alrw2}---a computationally efficient version that requires data-splitting---to compute their p-values. Instead of using \eqref{eq:p-value}, a \emph{nonconformity function} $A:\ZZ\to\overline{\mathbb{R}}$ is determined from the training set (usually by training an ML model). Computing p-values requires in addition a separate \emph{calibration set}. The p-values are then computed by
\begin{equation}\label{eq:inductive-p-value}
	p_n = \tfrac{|\{i=1,\dots,n:\alpha_i > \alpha_n\}| + \tau_n|\{i=1,\dots,n:\alpha_i=\alpha_n\}|}{n},
\end{equation}
where $\alpha_i = A(z_i)$, $z_1,\dots,z_n$ is the calibration set, and $\tau_n$ is again an independent random number distributed uniformly on the unit interval. The difference of \eqref{eq:inductive-p-value} from \eqref{eq:p-value} is that the \emph{nonconformity scores} $\alpha_i$ are computed using the same fixed function in the inductive case.

In addition to the Ville procedure, which raises an alarm when the martingale exceeds a threshold $c$, they also considered the conformal \emph{CUSUM} and \emph{Shirevev-Roberts} (SR) procedure \citep{vovk2021retrain,alrw2}; both of which compute statistics based on martingale trajectories to increase the robustness to capital loss in the early test stage. The details of these methods are beyond the scope of this paper, but we remark that because the CUSUM and SR statistics do not form proper martingales, Ville's inequality does not hold, which means that they gain robustness at the cost of exact false-alarm control.

The Ville procedure, outlined in \cite{vovk2021retrain}, to decide when to retrain a model is briefly the following:
\begin{enumerate}
	\item
	      Randomly split the training set into three folds of equal sizes. The random shuffling ensures that the training set is exchangeable.
	\item
	      Use each fold in turn as the calibration set and use the two remaining fold as the \emph{proper training set}.
	\item
	      For each fold, train an ML model on the proper training set, use the calibration set to compute p-values for the calibration set and the test set (in that order) in the semi-online mode where the calibration set is extended to include each new observation. Then run a CTM on the p-values.
\end{enumerate}
This way, we obtain three martingale trajectories; one for each fold. If any of the martingales exceeds the threshold, e.g. 100, we raise a warning saying that the model should be retrained. For the threshold 100, the probability of ever raising a false alarm is at most 3\%.

Following \cite{vovk2021retrain}, we considered two scenarios. In scenario 0, our test set consisted of 1599 randomly chosen white wines, which is referred to as \emph{test set 0}. All 1599 red wines (randomly permuted) for \emph{test set 1} are the focus of scenario 1. We normalised the data using \texttt{StandardScaler} from \texttt{scikit-learn} \citet{scikit-learn} by fitting the scaler on the training set, which consists of the remaining 3299 white wines. Our conformity function was
\begin{equation}
	A(z) = A(x,y) = |y - \hat{y}|
\end{equation}
where $\hat{y}$ is the prediction for $y$ output by a Random Forest with default parameters from \texttt{scikit-learn}.

The training set was then randomly split into three folds of sizes 1100, 1100, and 1099. For both scenario 0 and scenario 1, we proceeded as outlined above, computing one martingale trajectory for each fold. This procedure was repeated for 1000 independent random seeds, resulting in 3000 martingale trajectories for each scenario. We considered two CTMs. We considered two CTMs: the Simple Legendre Jumper of order one (SLJ(k=1)) and the Variational Legendre Jumper with orders one and two (SLJ([1,2])). Our results for SLJ(k=1) are not directly comparable to \cite{vovk2021retrain} as they used a smaller grid and a different train/test split for their Monte Carlo study, which also included more ML algorithms than our reduced experiment.

\begin{figure}[ht]
	\centering
	\begin{subfigure}[b]{0.49\textwidth}
		\centering
		\includegraphics[width=\textwidth]{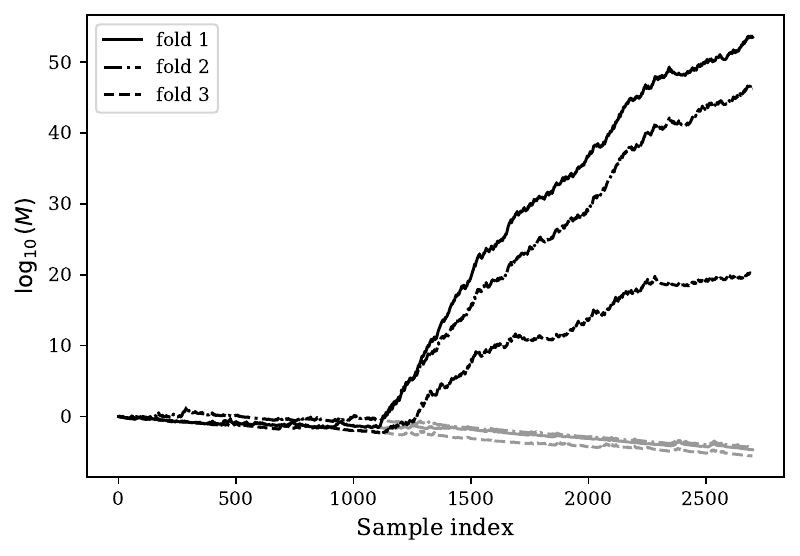}
		\caption{SLJ(k=1)}
		\label{subfig:SLJ}
	\end{subfigure}
	\begin{subfigure}[b]{0.49\textwidth}
		\centering
		\includegraphics[width=\textwidth]{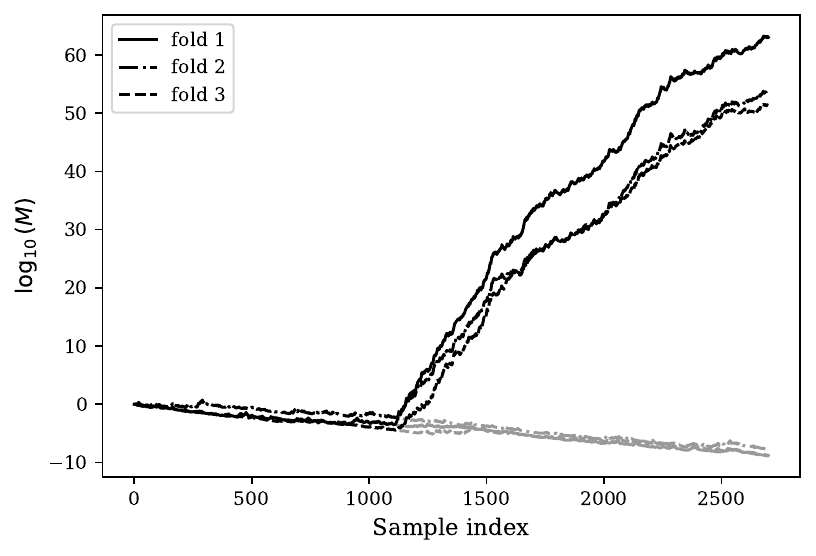}
		\caption{VLJ([1,2])}
		\label{subfig:VLJ}
	\end{subfigure}
	\caption{Martingale trajectories for scenario 0 (gray) and 1 (black).}
	\label{fig:retrain-plot}
\end{figure}

Figure \ref{fig:retrain-plot} illustrates the martingale trajectories for each fold in both scenarios. In scenario 0, there is no change point. Consequently, all martingale trajectories decrease, and no false alarm is raised. In scenario 1, the test set is from a different distribution than the training set, which means there is a change point at $n=1100$ (or $n=1099$ in case of the slightly smaller fold). All trajectories grow rapidly after the change point, but VLJ([1,2]) grows more consistently and reaches significantly larger values because it is not limited to betting only on the change in mean, but also the variance.

The median detection delays, defined as the ordinal number of the observation in the test set at which the alarm occurs, together with the interquartile intervals for the delays (the intervals whose endpoints are the lower and upper quartiles), with thresholds $c=10^2,10^4,10^6$, are listed in Table \ref{tab:detection_delay_grouped}. VLJ([1,2]) detects the change point faster than SLJ(k=1) for all thresholds $c$, and the improvement increases with $c$. Thus, the Variational Legendre Jumper reduces the median delay to signal an out-of-distribution shift compared to the classical Simple Jumper. This faster detection translates directly to fewer corrupt inferences passing through a live production pipeline before the automated retraining triggers.
\begin{table}[ht]
	\centering
	\begin{tabular}{ccccc}
		                   & \multicolumn{2}{c}{\textbf{Median Delay [IQR]}} & \multicolumn{2}{c}{\textbf{VLJ Improvement}}                                              \\
		\cmidrule(lr){2-3} \cmidrule(lr){4-5}
		\textbf{Threshold} & SLJ ($k=1$)                                     & VLJ ($[1,2]$)                                & \textbf{Absolute} & \textbf{Relative (\%)} \\
		\midrule
		$10^2$             & 86 [70, 104]                                    & 83 [68, 104]                                 & $-$3              & 3.5\%                  \\
		$10^4$             & 125 [104, 149]                                  & 112 [93, 134]                                & $-$13             & 10.4\%                 \\
		$10^6$             & 165 [140, 193]                                  & 141 [117, 168]                               & $-$24             & 14.5\%                 \\
	\end{tabular}
	\caption{Median detection delay and and improvement of VLJ([1,2]) over SLJ(k=1) (equivalent to the Simple Jumper).}
	\label{tab:detection_delay_grouped}
\end{table}

\begin{table}[ht]
	\centering
	\begin{tabular}{c|c}
		Scenario  & Median MAD [IQR]      \\
		\hline
		0 (white) & 0.486 [0.479, 0.493]  \\
		1 (red)   & 0.814 [ 0.776, 0.857]
	\end{tabular}
	\caption{Median and interquartile range for the MAD in each scenario.}
	\label{tab:MAD}
\end{table}

To illustrate the practical consequences of not retraining, we also report the mean absolute deviation (MAD) of the Random Forest model over the 3000 folds in each scenario. Because the distribution of the test set is different from that of the training set in Scenario 1, where we trained on white wines and predicted the quality of red wines, the MAD was significantly worse. The Variational Legendre Jumper was able to detect the change point (located at the start of test set 1) faster by betting on higher moment deviation from uniformity on the p-values, which translates to fewer low-quality predictions, caused by the distribution shift, before we can take action to mitigate the harm.

\section{Concluding discussion}\label{sec:conclusion}
We have introduced a family of conformal test martingales based on the shifted Legendre polynomials. The Simple Legendre Jumper generalises the well-known Simple Jumper by replacing the linear betting function with an arbitrary-degree shifted Legendre polynomial, enabling detection of deviations in variance, skewness, and higher-order moments individually. The Product Legendre Jumper combines multiple polynomial degrees into a single product betting function, allowing the martingale to simultaneously exploit shifts across several moments. To overcome the exponential state-space scaling inherent in the product construction---what we have termed the jumping tax---the Variational Legendre Jumper factorises the joint adaptation problem into independent marginal sub-jumpers via a mean-field approximation, reducing the per-step cost from $O(g^{|K|})$ to $O(|K|g)$.

An empirical evaluation of the Wine Quality dataset demonstrated that the combined methods (PLJ and VLJ) consistently outperformed any single-degree SLJ under real-world distributional shifts, whereas the variational approximation accumulated capital nearly identical to that of the exact product at a fraction of the computational cost. Importantly, all methods remained well-behaved under exchangeability, losing only a modest amount of capital to the jumping mechanism. As shown in Section~\ref{sec:empirical}, this loss was effectively eliminated by the Composite Legendre Jumper, which mixes over several jumping rates in direct analogy with the standard Composite Jumper \citep{alrw2}. The composite construction provides both a wealth floor and robustness to rate misspecification, making it the natural default for practitioners.

\medskip

Although the Variational Legendre Jumper empirically accumulates capital at a rate nearly identical to that of its exact product counterpart, the theoretical bounds of the mean-field approximation remain an open problem. In particular, deriving formal error bounds on the loss of statistical power incurred by factorising the joint adaptation would rigorously quantify the trade-off between computational efficiency and exactness. Establishing such bounds, perhaps by adapting divergence techniques from variational inference theory \citep{wainwright2008graphical},would address a key theoretical gap and provide explicit convergence guarantees for polynomial spaces with $|K| \ge 3$, where the approximation quality is currently uncharacterised.

The choice of polynomial degrees in $K$ connects to a long-standing principle in classical hypothesis testing. Historically, the standard advice for Neyman's smooth test is to choose a small, fixed integer $k$ and include all degrees from one to $k$, most commonly between 2 and 4. Neyman originally recommended $k=4$, although he was careful to point out that ``My personal feeling is that in most practical cases, there will be no need to go beyond the fourth order test. But this is only an \emph{opinion} and not any mathematical result'' \citep[Page 194]{neyman1937smooth}. In our single empirical evaluation on the Wine Quality dataset, $K=\{1,2\}$ captures substantial gains over $K=\{1\}$, whereas $K=\{1,2,3\}$ provides only moderate additional improvements on most orderings, although larger gains appear under the original data order. This tentatively aligns with the classical principle that low-degree polynomials suffice for detecting distributional shifts, although broader validation across diverse datasets and shift types is needed to assess the generality of this observation.

Within the Legendre framework, an adaptive or data-driven mechanism for selecting the polynomial degrees to include in $K$ could further improve efficiency by avoiding unnecessary orders. This direction is motivated by Ledwina's approach to Neyman's smooth test, which employs Schwarz's Bayesian Information Criterion (BIC) to automatically select the number of components rather than fixing the dimension in advance \citep{ledwina1994datadriven,schwarz1978estimating}. Ledwina notes that the wrong choice of degree results in a power loss ranging from 21\% to 48\%, underscoring the importance of principled degree selection. However, our sequential martingale framework differs fundamentally from the batch-based exponential family setting: our betting functions use Legendre polynomials directly, rather than embedding them within a full likelihood framework. Consequently, adapting the BIC-based selection directly would not be appropriate. Developing a data-driven degree selector suitable for the online conformal setting, perhaps through sequential model comparison or empirical validation analogous to our composite construction, remains a valuable avenue for future work.

\medskip

Our demonstration was limited to a single dataset and rather simple nonconformity measures. Future work could evaluate the various Legendre Jumpers on several real-world datasets using various nonconformity measures. Moreover, the Ville procedure is not the only mode of conformal testing: the conformal framework also includes the conformal CUSUM and Shiryaev--Roberts procedures for multistage testing \citep{vovk2021retrain,alrw2}. We hope to investigate the performance of Legendre Jumpers in these procedures, including a comparison with other conformal test martingales. Additionally, the jumping martingale framework can be extended naturally to sleeping and waking mechanisms \citep{Bostrom2026,alrw2}, which can be combined with higher-degree betting functions. Finally, a theoretical analysis of the growth rate of Legendre Jumper martingales under specific parametric alternatives would complement the empirical findings presented in this study.

\medskip

While our focus has been on rapid detection of distributional shifts, the practical utility of these methods extends beyond signalling when a problem occurs. A complete deployed system must address a critical three-stage pipeline: (i) detecting the shift, (ii) deciding to retrain, and (iii) handle predictions during the retraining phase. In Section \ref{sec:empirical}, we illustrate that including higher-order Legendre polynomials in the betting function can be beneficial for improving detection delay. However, we did not discuss the actions that should be triggered by the alarm, which part of the available data to use for retraining, and how to handle predictions during the time it takes to retrain. A natural solution to the first question, mentioned briefly by Vovk et al. \citep{vovk2021retrain}, is to run a conformal test martingale backward (trained on recent data) until it raises a warning. The data before the warning point (or really after, since we run the martingale backward) could be used to retrain the model. The second question becomes particularly important if the training process is lengthy. In such cases, conformal test martingales can be used to protect the predictions of ML models \citep{vovk2021protectedprobabilisticclassification,vovk2021enhancementpredictionalgorithmsbetting,alrw2}. The betting functions of the martingale are then used to transform the predictions of the model with provable calibration improvements if the martingale grows, ensuring that the prediction quality during retraining do no suffer too much. Investigating how the Legendre Jumper martingales perform on both detection and protection tasks, as well as their integration within an end-to-end deployment pipeline, appears to be an interesting and practically important subject for future work.

We note that the choice of Legendre polynomials is canonical: they are the unique orthogonal system with respect to the uniform measure on $[0, 1]$, matching the null distribution of conformal p-values. Other polynomial families (e.g., Chebyshev and Jacobi) are orthogonal under non-uniform weights and therefore do not satisfy the integral constraint $\int_0^1 f(p) dp = 1$ without modification; exploring such weighted alternatives would correspond to testing against non-uniform nulls, a problem we did not pursue here.

\section*{Acknowledgement}
I am grateful to Alexander Gammerman for valuable comments on an early draft of this paper that inspired me to reframe the introduction and add the Monte Carlo study in Section \ref{subsec:retrain-or-not}.



\printbibliography
\end{document}